\documentclass[manuscript]{jmlr} % W&CP article
\usepackage{booktabs}
\usepackage{multirow}
\usepackage[load-configurations=version-1]{siunitx} % newer version
\theorembodyfont{\upshape}
\theoremheaderfont{\scshape}
\theorempostheader{:}
\theoremsep{\newline}

\usepackage[font=small,skip=0pt]{caption}
\vspace{-35pt}
\title[Iterative Amortized Bayesian Inverse Problems]{Refining Amortized Posterior Approximations using Gradient-Based Summary Statistics}
\vspace{-10pt}

 \author{\Name{Rafael Orozco} \Email{rorozco@gatech.edu}\\
 \addr Georgia Institude of Technology
 \AND
 \Name{Ali Siahkoohi} \Email{alisk@rice.edu}\\
 \addr Rice University
  \AND
 \Name{Mathias Louboutin} \Email{mlouboutin3@gatech.edu}\\
 \addr Georgia Institute of Technology
  \AND
 \Name{Felix J. Herrmann} \Email{felix.herrmann@gatech.edu}\\
 \addr Georgia Institute of Technology
 }

\begin{document}
\vspace{-65pt}
\maketitle
\thispagestyle{empty}

\vspace{-35pt}
\begin{abstract}
%This is the abstract for this article. It is optional for AABI submissions.

% We present an iterative framework for enhancing amortized approximations of Bayesian posteriors in inverse problems, inspired by loop-unrolled gradient descent and theoretically grounded in maximally informative summary statistics. Our method extends amortized variational inference, which is constrained by the expressive family of distributions and training data, leading to approximation errors like the \textit{amortization gap}. We summarize observations using a gradient-based summary statistic and employ conditional normalizing flows to learn a probabilistic update of the unknown parameter based on the summary statistic. This results in iterative refinement of the amortized approximation without extra training data. We validate our method using a Gaussian problem, showing improved posterior approximation with each iteration.\footnote{Code to reproduce experiments included as anonymized source code. Github repo to be released once deanonymized.}
% %\footnote{Code to reproduce experiments can be found at https://github.com/slimgroup/IterativeAmortizedPosterior.jl.}
% Applying our approach to transcranial ultrasound, a high-dimensional nonlinear inverse problem governed by wave physics, we demonstrate enhanced posterior quality through better image reconstruction with the posterior mean.
% %and uncertainty calibration compared to non-iterative amortized inference.

We present an iterative framework to improve the amortized approximations of posterior distributions in the context of Bayesian inverse problems, which is inspired by loop-unrolled gradient descent methods and is theoretically grounded in maximally informative summary statistics. Amortized variational inference is restricted by the expressive power of the chosen variational distribution and the availability of training data in the form of joint data and parameter samples, which often lead to approximation errors such as the \textit{amortization gap}. To address this issue, we propose an iterative framework that refines the current amortized posterior approximation at each step. Our approach involves alternating between two steps: (1) constructing a training dataset consisting of pairs of summarized data residuals and parameters, where the summarized data residual is generated using a gradient-based summary statistic, and (2) training a conditional generative model---a normalizing flow in our examples---on this dataset to obtain a probabilistic update of the unknown parameter. This procedure leads to iterative refinement of the amortized posterior approximations without the need for extra training data. We validate our method in a controlled setting by applying it to a stylized problem, and observe improved posterior approximations with each iteration. Additionally, we showcase the capability of our method in tackling realistically sized problems by applying it to transcranial ultrasound, a high-dimensional, nonlinear inverse problem governed by wave physics, and observe enhanced posterior quality through better image reconstruction with the posterior mean.\footnote{Code to reproduce experiments included as anonymized source code. Github repository to be released after the review process.}
%\footnote{Code to reproduce experiments can be found at https://github.com/slimgroup/IterativeAmortizedPosterior.jl.}
%and uncertainty calibration compared to non-iterative amortized inference.

\end{abstract}

% Keywords may be removed
%\begin{keywords}
%List of keywords
%\end{keywords}

\section{Introduction}
\label{sec:intro}
We aim to solve Bayesian inverse problems where indirect observations $\mathbf{y}$ of parameters $\mathbf{x}$ are given by the forward operator $\mathcal{F}$ and noise $\pmb\varepsilon$:

\begin{equation} \label{eq:forward}
\mathbf{y} = \mathcal{F}(\mathbf{x}) + \pmb\varepsilon.
\end{equation}
%ACTUALLY DOESNT NEED TO BE ADDITIVE NOISE. WE JUST NEED THE SCORE. 

Our goal is to estimate the posterior distribution $p(\mathbf{x}\mid \mathbf{y})$ of parameters $\mathbf{x}$ given observation $\mathbf{y}$. Bayes' theorem relates the posterior to the likelihood function and prior distribution as: $p(\mathbf{x}\mid \mathbf{y}) \propto p(\mathbf{y}|\mathbf{x})p(\mathbf{x})$ where the likelihood $p(\mathbf{y}|\mathbf{x})$ depends on the noise distribution $p(\pmb\varepsilon)$ and the forward operator $\mathcal{F}$. 
%Challenges in this problem include high-dimensionality requiring numerous posterior samples to calculate statistics, non-linear forward operator $\mathcal{F}$ leading to multi-modal posteriors, and obtaining analytical expressions for non-Gaussian priors. 
Challenges in this problem include: (1) high-dimensionality of unknown parameters, which necessitates the computation of numerous posterior samples to calculate statistics; (2) a non-linear forward operator $\mathcal{F}$, resulting in multi-modal posteriors; and (3) obtaining analytical expressions for non-Gaussian priors that capture our prior knowledge about the parameters. To overcome these challenges, recent work has exploited deep neural networks by learning the high-dimensional and non-linear behaviour of these inverse problems and by approximating unknown prior distributions from examples. These methods typically fall under the umbrella of variational inference (VI) since the computational complexity of posterior sampling is exchanged for neural network optimization \cite{jordan1999introduction}. There are two VI approaches: amortized VI and non-amortized VI \cite{zhang20233}. 

%\ali{could we throw in some refs for amortized and non-amortized?}
Amortized VI can leverage deep generative models to approximate posterior distributions, offering wide applicability and online inference efficiency \cite{radev2020bayesflow}. %CITE OUR WORK. 
By online efficiency, we mean that amortized VI methods entail a computationally expensive pre-training phase (offline cost) but its cost at inference (online cost) is low because after training they can cheaply compute posterior inference results  
for many different observations $\mathbf{y}$ drawn from the same distribution as training data.
In contrast, non-amortized VI does not have an offline cost thus all the computational cost is paid online at inference time by solving an optimization problem related to the potentially expensive operator $\mathcal{F}$ and its gradient.  Non-amortized inference results can not be reused so the optimization will need to be repeated for every new observation. Since non-amortized VI is specialized to a single observation $\mathbf{y}$ its performance is typically better than its amortized counterpart since amortized VI averages performance over many observations; a phenomena denoted as the amortization gap \cite{marino2018iterative}. Our work aims to bridge the gap with principled additional offline computation, preserving the fast online inference properties of amortized VI while achieving performance closer to non-amortized VI approaches. 
%\ali{There is a subtlety missing here, and that is the learned prior in AVI. The non-amortized does not have that unless it incorporates a learned prior. It might be useful to mention this somewhere above as a disadvantage of non-amortized as well since at the moment only the cost of the process is mentioned, and our iterative method also involves (though less) evaluations of the forward operator.} \rafael{very true. will add if we have space}
%I WOULD LIKE AN IMAGE HERE. Maybe a SCHEMATIC OF AMORTIZATIN GAP AND THEN OUR METHOD DURING INFERENCE . \ali{great idea, basically can even go  on the first page under the abstract and introduction can start in page 2} 
\vspace{-9pt}
\section{Related work}
\label{sec:related}
\begin{figure}[htbp]
\floatconts
  {fig:gaussianprob}
  {\caption{Our method iteratively improves its posterior approximation as compared to the analytical posterior.}}
  {%
   \subfigure[Posterior mean][b]{\label{fig:gaussianmean}%
      \includegraphics[width=0.33\linewidth]{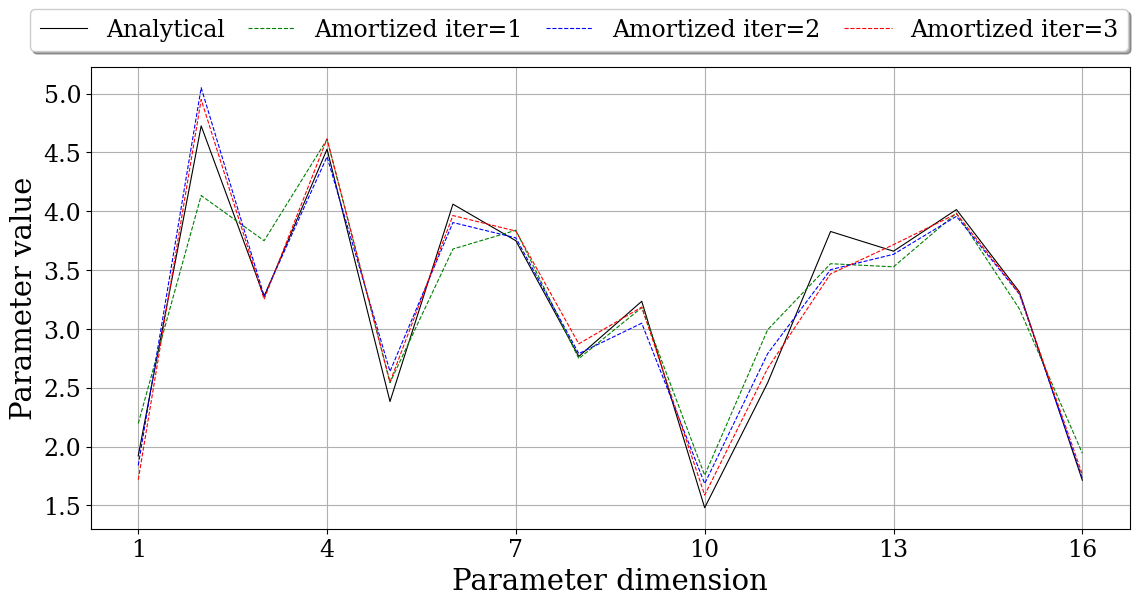}} 
    \subfigure[Posterior covariance][b]{\label{fig:gaussiancov}%
      \includegraphics[width=0.65\linewidth]{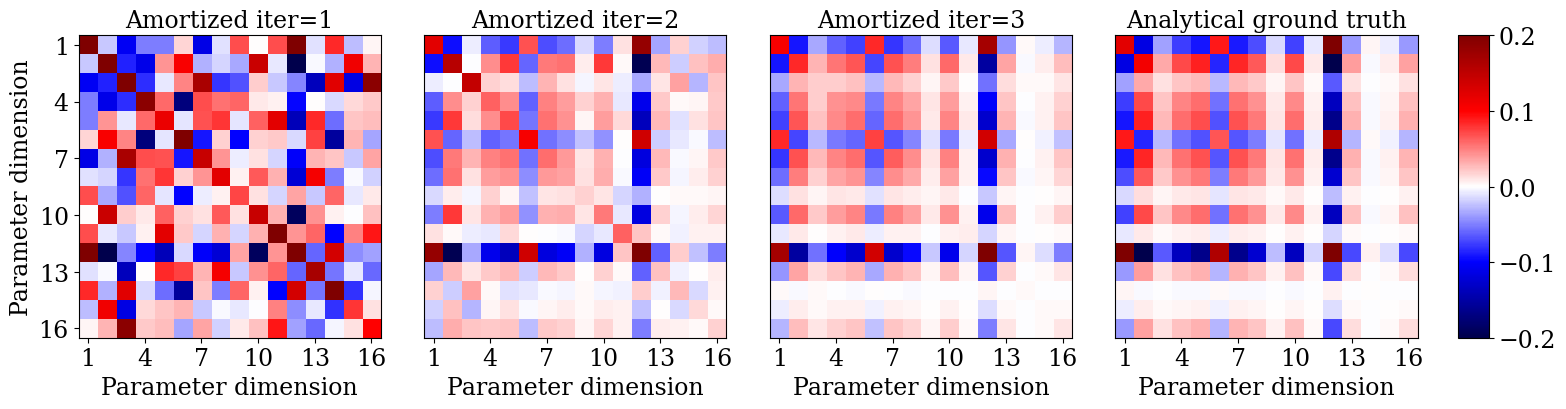}}%
  }
\end{figure}

 Our method is related to the two-step process in \cite{siahkoohi2021preconditioned} starting with amortized VI then switching to non-amortized VI taking advantage of the implicitly learned prior. \cite{siahkoohi2022reliable} also performs an amortized then non-amortized step with the goal of correcting for distribution changes from the training data.
 %The distribution change is described as a shift in the latent space of the normalizing flow that can be corrected efficiently. 
 Bayesian results in ultrasound imaging were previously shown \cite{bates2022probabilistic} but used a mean field approximation constraining it to point wise variances. The method we propose is not limited to Gaussian priors and recovers the full posterior covariance \figureref{fig:gaussiancov}. Deep-GEM \cite{gao2021deepgem} is similar to our work since it iterates between normalizing flow training and gradient-based optimization. In contrast with Deep-GEM, our method trains an amortized normalizing flow, and at each iteration only one gradient step is needed; leading to an amortized procedure with efficient online inference. 
\section{Methods }
\label{sec:methods}
\subsection{Amortized variational inference}
\label{sec:avi}
Reducing the costs of Bayesian inference in %high-dimensional non-linear
inverse problems with computationally costly forward operators can be achieved via amortized VI \cite{radev2020bayesflow}. This is implemented by learning a parametric conditional distribution $p_{\theta}(\mathbf{x}\mid \mathbf{y})$ (here we will use normalizing flows \cite{dinh2016density}) that approximates the posterior for a distribution of observations. The approximate posterior quality is measured by the Kullbeck-Leibler (KL) divergence calculated in expectation (amortized) over the distribution of observations $p(\mathbf{y})$. Amortized VI minimizes this expected KL divergence  with respect to $\theta$ 
%\boldsymbol{\theta}$:
% \begin{equation}
% \hat\theta = \underset{\mathbf{\theta} }{\operatorname{arg \, min}} 
%  \,\mathbb{E}_{p(\mathbf{y})} \Bigl[ \mathbb{KL} \left( \,  p(\mathbf{x}\mid \mathbf{y}) \, \,||\,\, p_{\theta}(\mathbf{x}\mid \mathbf{y}) \right) \Bigr]. 
% \label{eq:KL}
% \end{equation}
\begin{align}
   &\underset{\mathbf{\theta} }{\operatorname{min}} 
 \,\mathbb{E}_{p(\mathbf{y})} \Bigl[ \mathbb{KL} \left( \,  p(\mathbf{x}\mid \mathbf{y}) \, \,||\,\, p_{\theta}(\mathbf{x}\mid \mathbf{y}) \right) \Bigr] 
 \nonumber 
   = \,\underset{\mathbf{\theta} }{\operatorname{min}} 
 \,\mathbb{E}_{p(\mathbf{y})}  \mathbb{E}_{p(\mathbf{x}\mid \mathbf{y})} \Bigl[-\log p_{\theta}(\mathbf{x}\mid \mathbf{y})  +  p(\mathbf{x}\mid \mathbf{y})  \Bigr] \nonumber \\ 
 = \,&\underset{\mathbf{\theta} }{\operatorname{min}} 
 \,\mathbb{E}_{p(\mathbf{x}, \mathbf{y})} \Bigl[-\log p_{\theta}(\mathbf{x}\mid \mathbf{y})  +  p(\mathbf{x}\mid \mathbf{y})  \Bigr]  \label{eq:KL} 
\end{align}
where we have rewritten the terms to avoid an expectation over the posterior distribution in favor of an expectation over the joint distribution $p(\mathbf{x},\mathbf{y})$. %\ali{In addition, the term $p(\mathbf{x}\mid \mathbf{y})$ in the last line of the equation above can be dropped as it is constant with respect to $\boldsymbol{\theta}$.}
In practice, a dataset of $N$ training examples $\{({\mathbf{x}^{(n)},\mathbf{y}^{(n)}})\}_{n=1}^N$ from the joint distribution is used to approximate the expectation. For conditional normalizing flows (CNF), the above optimization problem simplifies to: 
\begin{equation} \label{eq:train-cond}
\underset{\theta}{\operatorname{ min}}
 \,  \frac{1}{N} \sum_{n=1}^{N}  \left( \lVert  f_{\theta}(\mathbf{x}^{(n)};\mathbf{y}^{(n)}) \rVert_{2}^2 - \log \left|  \det{  \mathbf{J}_{f_{\theta}}} \right| \right),
\end{equation} 
%Our method can be implemented with any conditional distribution learning approach (conditional GAN, VAE, diffusion) but we concentrate on normalizing flows since their invertibility allows for scaling to high-dimensional parameters while keeping training memory requirements low. \ali{This can indeed be highlighted (even in the abstract and introduction), but right here as the first sentence under the AVI section seems out of place.} 
where $f_{\theta}$ is a CNF implemented as in \cite{ardizzone2019conditional} and $\mathbf{J}_{f_{\theta}}$ is the Jacobian of the CNF. The success of the CNF in approximating the posterior depends on the amount of training data used to estimate the expectation in \equationref{eq:KL}. For high-dimensional inverse problems, limited training data leads to poor expectation estimates and therefore poor posterior approximations. To improve the accuracy of this posterior approximation, we will discuss the role of summary statistics in high-dimensional inference and how they ultimately lead to a iterative formulation for improving the posterior approximation without additional training data. 

\subsection{Summary statistics for high-dimensional inverse problems}
%\ali{Please use the "Felix structure" as exemplified above. The following paragraph suddenly appears out of nowhere. As a reader, I do not understand why I am required to read about summarizing data and fiducial points. Can we discuss summarizing the data in the previous section as a means to deal with high-dimensionality? We then assert that not only do we lack sufficient data, but a simple adjoint is insufficient to make summarize the data. Consequently, the reader is interested in learning more about how to obtain a better summary of the data.}
Efficient inference procedures are difficult to implement when observations $\mathbf{y}$ are high-dimensional \cite{cranmer2020frontier}. Thus efficient inference procedures typically use a summary statistic to summarize large observations.
\cite{alsing2018generalized} demonstrated that $\mathbf{y}$ can be summarized by the score of the log-likelihood at a fiducial $\mathbf{ x}_0$. In this context, the fiducial $\mathbf{x}_0$ refers to a reliable estimate of the unknown $\mathbf{x}$. Furthermore, it has been proven that this score is maximally informative \cite{deans2002maximally}, as determined by the Fisher information it contains needed to infer $\mathbf{x}$. 
The score summary statistic is defined as
\begin{equation} \label{eq:score}
\mathbf{\bar y} := \nabla_{\mathbf{x}_0} \log p(\mathbf{y} | \mathbf{x}_0).
\end{equation} 
The score $\mathbf{\bar y}$ is maximally informative specifically around the fiducial $\mathbf{x}_0$ so approximate posterior learning with a score summary statistic has a modified implementation. First we change the target to be the update $\mathbf{\Delta x} =  \mathbf{x} - \mathbf{x}_0$ and using \equationref{eq:train-cond} we learn the conditional distribution $p(\Delta \mathbf{x}|\mathbf{\bar y})$, i.e., 
\begin{equation}
  \begin{aligned}
    &\hat\theta = \underset{\theta}{\operatorname{arg \, min}}
    \,  \frac{1}{N} \sum_{n=1}^{N} \Biggl( \lVert  f_{\theta}(\mathbf{\Delta x}^{(n)};\mathbf{\bar y}^{(n)}) \rVert_{2}^2 - \log \left|  \det{  \mathbf{J}_{f_{\theta}}} \right| \Biggr) \label{eq:train-cond-summary}. \\
    %&\text{where } \mathbf{\Delta x}^{(n)} =  \mathbf{x}^{(n)} - \mathbf{x}_0^{(n)}. \nonumber \\
      %\mathbf{\bar y}^{(n)} &= \nabla_{\mathbf{x}_0} \log p(\mathbf{y}^{(n)} | \mathbf{x}_0^{(n)}).
  \end{aligned}
\end{equation}
Then the approximate posterior is sampled by adding the learned conditional distribution to the fiducial, i.e., $p(\mathbf{x}|\mathbf{y}) \approx \mathbf{x}_0 + p_{\hat \theta}(\mathbf{\Delta x} \mid \mathbf{\bar y})$, where the conditional samples from $p_{\hat \theta}(\mathbf{\Delta x} \mid \mathbf{\bar y})$ are generated by the CNF with optimized weights $\hat \theta$. 
%\ali{sorry I am lost why we are doing this.} 
Using the score as a summary in the context of Bayesian inference has been implemented \cite{adler2018deep} but because of finite training data and because of the amortization gap the amortized results can be lacking compared with non-amortized results \cite{siahkoohi2022reliable,orozco2023amortized}. 

\subsection{Iterative amortized variational inference}
To improve the amortized approximation, we take inspiration from loop-unrolled gradient descent 
%\ali{maybe add a few refs?}
and identify that the network trained in \equationref{eq:train-cond-summary} also takes the gradient as input (score is gradient of log-likelihood). Thus we suggest to interpret the fiducial $\mathbf{x}_0$ as the first in a sequence of $L$ updating iterations $\mathbf{x}_0, \mathbf{ x}_1 \ldots \mathbf{ x}_L$ where the CNF
has learned a probabilistic update to the fiducial $p_{\hat \theta_0}(\mathbf{\Delta x}_0 \mid \mathbf{\bar y}_0)$. We added subscript notation to emphasize that the weights $\hat \theta_0$, update $\mathbf{\Delta x}_0$ and score $\mathbf{\bar y}_0$ correspond to the $0$th fiducial. To update the fiducial, we use a single point estimate from the learned distribution $p_{\hat \theta}(\mathbf{\Delta x}_0 \mid \mathbf{\bar y}_0)$. In this work, we concentrate on the approximate posterior mean  $\mathbb{E}_{p_{\hat \theta_0}(\mathbf{\Delta x}_0 \mid \mathbf{\bar y}_0)}\, \left[\mathbf{\Delta x}_0 \mid \mathbf{\bar y}_0 \right] $: 
\begin{equation} \label{eq:update}
\mathbf{ x}_{1} = \mathbf{x}_0 + \mathbb{E}_{p_{\hat \theta_0}(\mathbf{\Delta x}_0 \mid \mathbf{\bar y}_0)}\, \left[\mathbf{\Delta x}_0 \mid \mathbf{\bar y}_0 \right],
  \end{equation} 
  where the expectation is estimated as in Appendix \equationref{eq:expectation}.
  We chose the posterior mean since it is the estimator with least variance \cite{adler2018deep}. The implications of using different point statistics 
  %such as the median or maximum a-posteriori are
  is left for future work. According to \cite{alsing2018generalized}, the informativeness of the score depends on how close the fiducial $\mathbf{ x}_0$ is to the ground truth $\mathbf{x}$. Therefore if the new fiducial $\mathbf{ x}_1$ is closer to $\mathbf{x}$ than $\mathbf{ x}_0$ then the score calculated at this new point $\mathbf{\bar y}_1$ will be a more informative summary statistic and lead to a better posterior approximation. We empirically verify this statement in \sectionref{sec:gaussiantoy}. 

By updating all fiducials $\mathbf{ x}_0^{(n)}$ in the training dataset with \equationref{eq:update}
and computing new scores $\mathbf{\bar y}_1^{(n)}$ we create a new dataset of pairs $\{({\mathbf{\Delta x}_1^{(n)}, \mathbf{\bar y}_1^{(n)}})\}_{n=1}^N$
that can be used to train a second CNF $f_{\hat \theta_1}$. In the algorithm we propose, this training and fiducial update procedure is continued alternated for $L$ iterations as outlined in Appendix \algorithmref{alg:train}.

At inference time, we take the trained networks $f_{\hat \theta_0}, f_{\hat \theta_1} \ldots , f_{\hat \theta_L}$ and the first fiducial $\mathbf{ x}_0$ then iterate on calculating scores $\mathbf{\bar y}_i$ at the $i$-th fiducial and updating the fiducial with the posterior mean until iteration $L-1$. The final posterior approximation is the last fiducial added to samples from the last learned conditional distribution. The full inference algorithm is shown in \algorithmref{alg:test}.

\begin{algorithm2e}
\caption{Inference phase}
\label{alg:test}
 % older versions of algorithm2e have \dontprintsemicolon instead
 % of the following:
 %\DontPrintSemicolon
 % older versions of algorithm2e have \linesnumbered instead of the
 % following:
 \LinesNumbered
\KwIn{Observation $\mathbf{y}$ and starting fiducial $\mathbf{x}_0$ }

\For{$i = 0$ \KwTo $L-1$}{
  %$\mathbf{\bar y} = \sum_{i=1}^{N_{s}}\mathbf{J}(\mathbf{x}_0,\mathbf{q}_i)^{\top}(\mathcal{F}(\mathbf{x}_0)\mathbf{q}_{i} - \mathbf{y^{}}_{i})$\;
  $\mathbf{\bar y}_{i} = \nabla_{\mathbf{ x}_i} \log p(\mathbf{y} | \mathbf{ x}_i)$\;
  $\mathbf{ x}_{i+1} = \mathbf{ x}_i +  \mathbb{E}_{p_{\hat \theta_i}(\mathbf{\Delta x}_i \mid \mathbf{\bar y}_i)}\, \left[\mathbf{\Delta x}_i \mid \mathbf{\bar y}_i \right] $\;
}
\KwOut{Final approximate posterior: $p(\mathbf{x}|\mathbf{y}) \approx p_{\hat \theta_L}(\mathbf{x}\mid\mathbf{ y}) = \mathbf{ x}_L + \, p_{\hat \theta_L}(\mathbf{\Delta x}_L\mid\mathbf{\bar y}_L)$ }
\end{algorithm2e}
\vspace{-19pt}
\section{Results }
\label{sec:results}
%\ali{increase the space before "Results"}
%\ali{Felix often suggests not to have nested sections without any text in between. I think this is a perfect place to put a preamble to the results section outlining what are the goals of each experiment (space permiting)}

\subsection{Validation with Gaussian example}
\label{sec:gaussiantoy}
Our proposed method is grounded in theoretical concepts but its application to our ultrasound medical imaging inverse problem is difficult to validate due to a non-linear forward operator and an unknown prior. To build trust, we first validate the our claims with an inverse problem that has an analytical posterior distribution to compare with. We solve an inverse problem with Gaussian prior on parameters and noise with a linear forward operator written as $\mathbf{y} =\mathbf{A}\mathbf{x}+\pmb\varepsilon$ where $\mathbf{A} \in\mathbb{R}^{64\times 16}$. Given an observation $\mathbf{y}$ there is an analytical expression for the posterior covariance and the posterior mean $\mathbb{E}_{p(\mathbf{ x} \mid \mathbf{ y})}\, \left[\mathbf{x} \mid \mathbf{ y} \right]$. See \cite{hagemann2021generalized} %\ali{nit picking: maybe the pattern recognition book instead?}
for derivation details. 
%We compare our approximate posterior with the analytical posterior. 

%\ali{If it isn't too much work it would help the readability if the lines were thicker and texts are larger in the above images.}

We trained our method using $N=1000$ training examples from the joint distribution and all starting fiducials $\mathbf{x}_0^{(n)}$ were set to zeros. In \figureref{fig:gaussianprob}, we observe that after each iteration, our posterior mean and covariance is a better approximation to the ground truth. In Appendix, \sectionref{apd:lowtrain}, we show that this trend is consistent by showing the average approximation error for a test set.

% \begin{figure}[htbp]
%  % Caption and label go in the first argument and the figure contents
%  % go in the second argument
% \floatconts
%   {fig:improveVI}
%   {\caption{Approximation errors due to small training datasets can be ameliorated with our iterativ method. RAFAEL: FIG IS PLACEHOLDER the posterior mean plot should have the same color scheme as the previous figure. }}
%   {\includegraphics[width=0.6\linewidth]{images/dim_data=80_dim_model=16_n_epochs=20_VI_quality_num_training.png}}
% \end{figure}

\subsection{High-dimensional ultrasound imaging}
\begin{figure}[htbp]
\floatconts
  {fig:medicalsetup}
  {\caption{Medical imaging inverse problem: (\textit{a}) 3D imaging hardware; (\textit{b}) 2D in-silico demo used in this work; (\textit{c}) Example starting fiducial $\mathbf{ x}_0$; (\textit{d}) Simulated observations $\mathbf{y}$;
 }}
  {%
    \subfigure[]{\label{fig:3dhelmet}%
    \raisebox{4.1mm}{
      \includegraphics[width=0.19\linewidth]{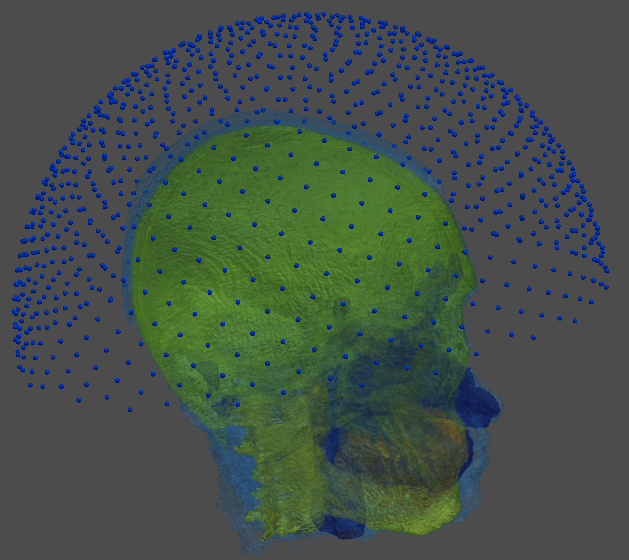}}}%
    \subfigure[]{\label{fig:2dsetup}%
      \includegraphics[width=0.25\linewidth]{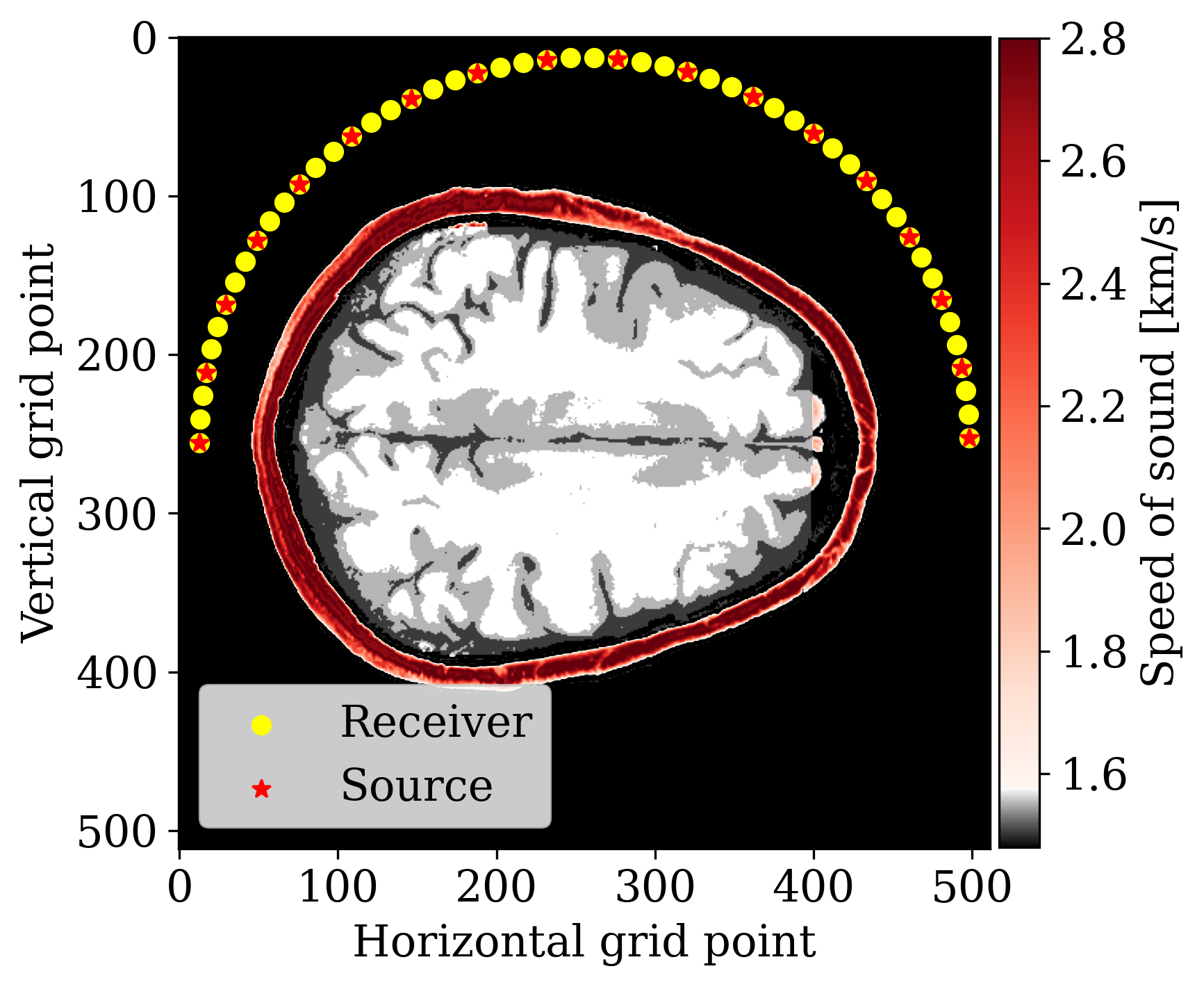}}
    \subfigure[]{\label{fig:startingbone}%
      \includegraphics[width=0.25\linewidth]{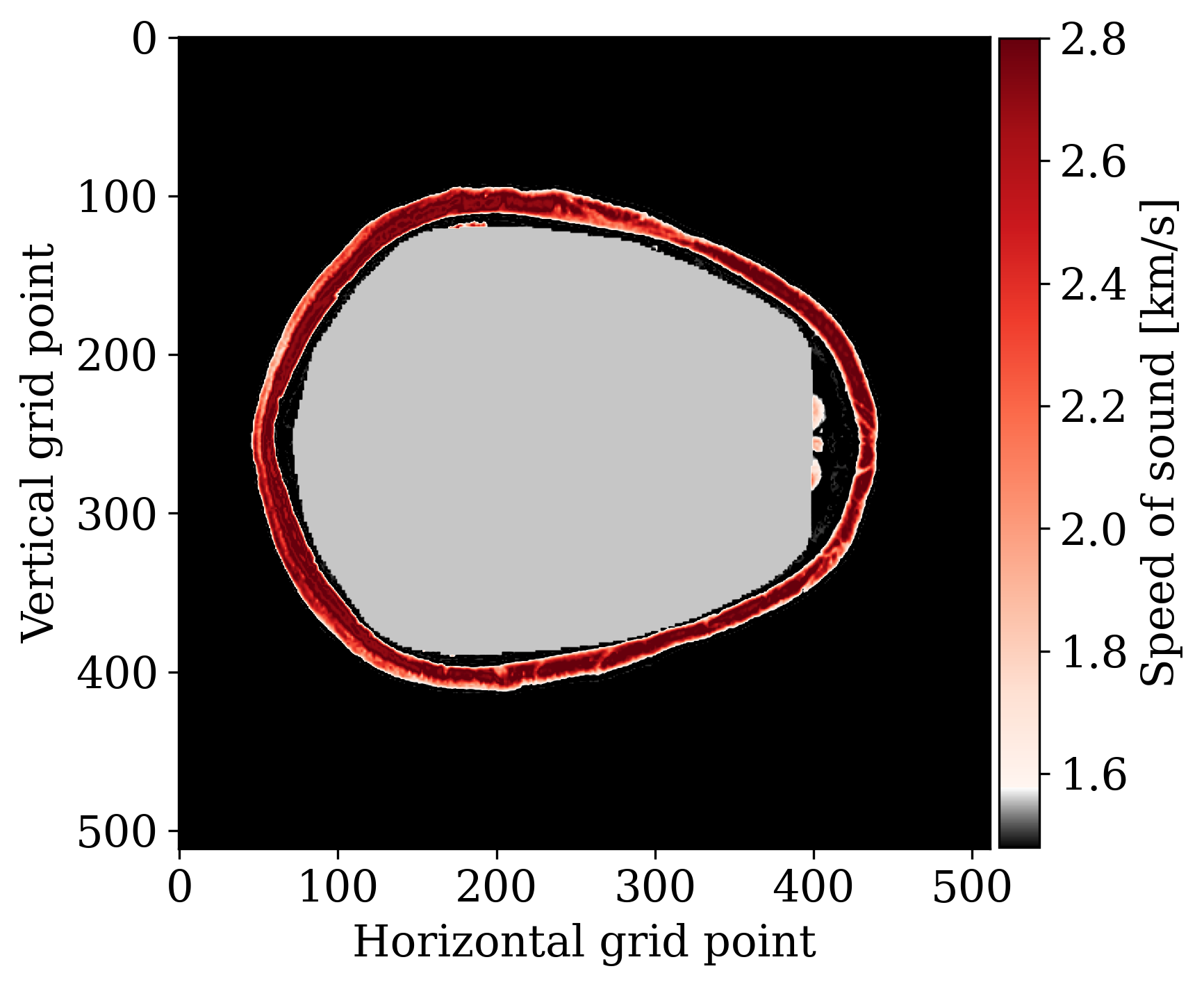}}
    \subfigure[]{\label{fig:acousticdata}%
     \raisebox{0.5mm}{
      \includegraphics[width=0.255\linewidth]{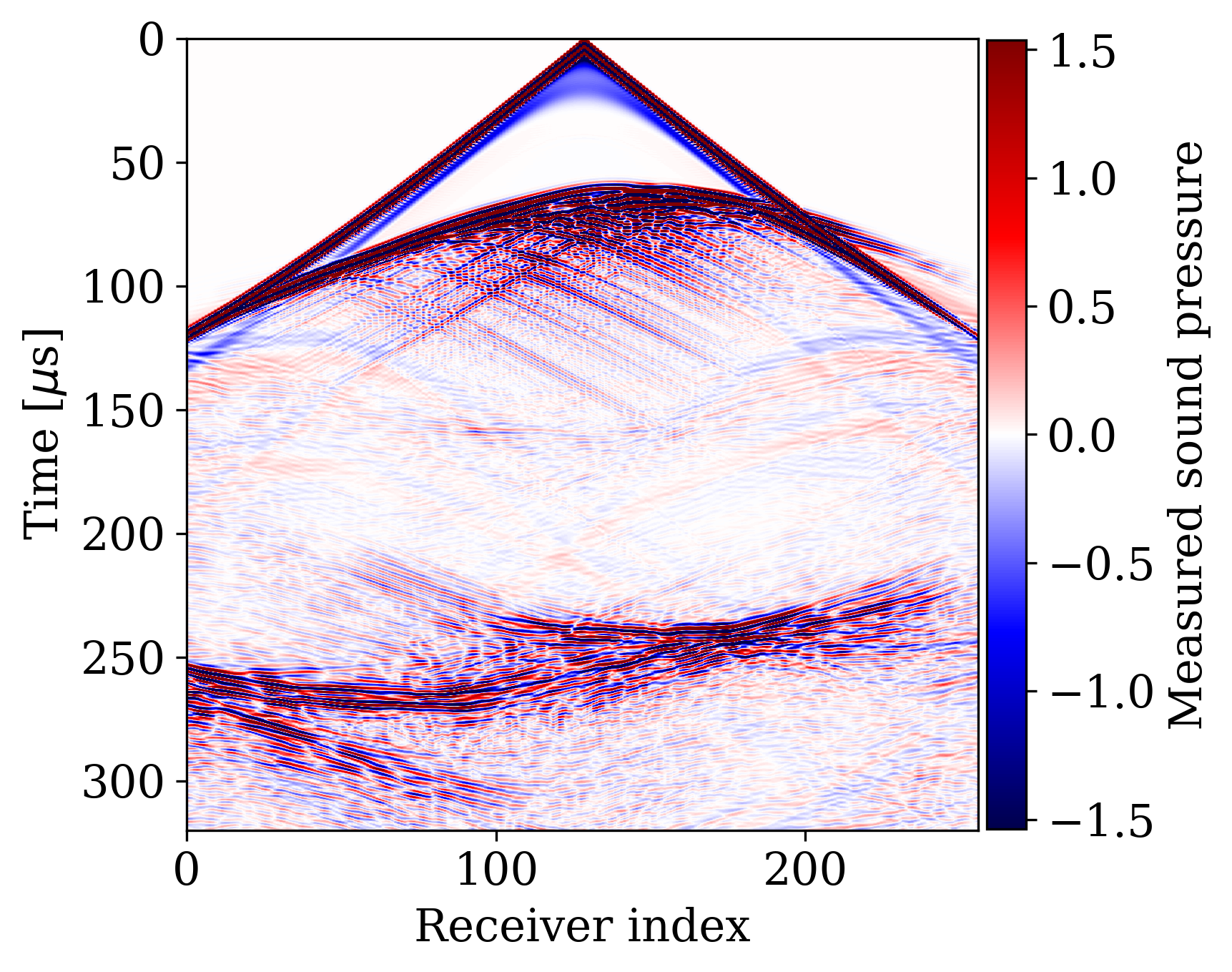}}}
  }
\end{figure}

We evaluate our iterative method on transcranial ultrasound computed tomography (TUCT). TUCT involves estimating speed of sound of brain tissue from measurements of ultrasound waves impinging on internal brain tissue. 
%\ali{if you feel a ref could be helpful please add} 
For a schematic of the hardware involved in the real-life 3D imaging setup see \figureref{fig:3dhelmet}. In this work, we consider a 2D slice \figureref{fig:2dsetup} of the 3D setup. A Bayesian framework is desirable for TUCT because the observations are noisy and the forward operator has a null space due to limited-view receivers, in other words: waves can only be transmitted and received from the top of the simulation \figureref{fig:2dsetup}. 
%\ali{in case fig 2a is not ours could we cite in the caption?} It is ours!
For details on TUCT, refer to \cite{guasch2020full, marty2021acoustoelastic}.

The TUCT inverse problem is high-dimensional ($512 \times 512$ grid here) and the forward operator $\mathcal{F}$ is non-linear and is characterized by the wave equation, which is parameterized by varying speed of sound and density. We use finite-differences software Devito \cite{devito-api} to solve the wave wave equation $\mathcal{F}$ and its Jacobian for scores/gradients. Our method requires samples from the prior $p(\mathbf{x})$ on brain speed of sound parameters that we generate by assigning acoustic values to the FASTMRI dataset \cite{zbontar2018fastmri}. We used MRI volumes from 250 subjects to make $N=2475$ training 2D slices. Due to pathological cycle skipping %\ali{, i.e., converging to undesired local minima,} 
in wave-based imaging \cite{virieux2009overview} the fiducial starts from the bone known (assumed to be calculated from X-ray as in \cite{aubry2003experimental}) and constant speed of sound for the internal tissue, refer to \figureref{fig:startingbone} for an example starting fiducial. 
\begin{figure}[htbp]
 % Caption and label go in the first argument and the figure contents
 % go in the second argument
\floatconts
  {fig:brainresult}
  {\caption{Our iterative method applied to a high-dimensional non-linear medical imaging problem. }}
  {\includegraphics[width=1\linewidth]{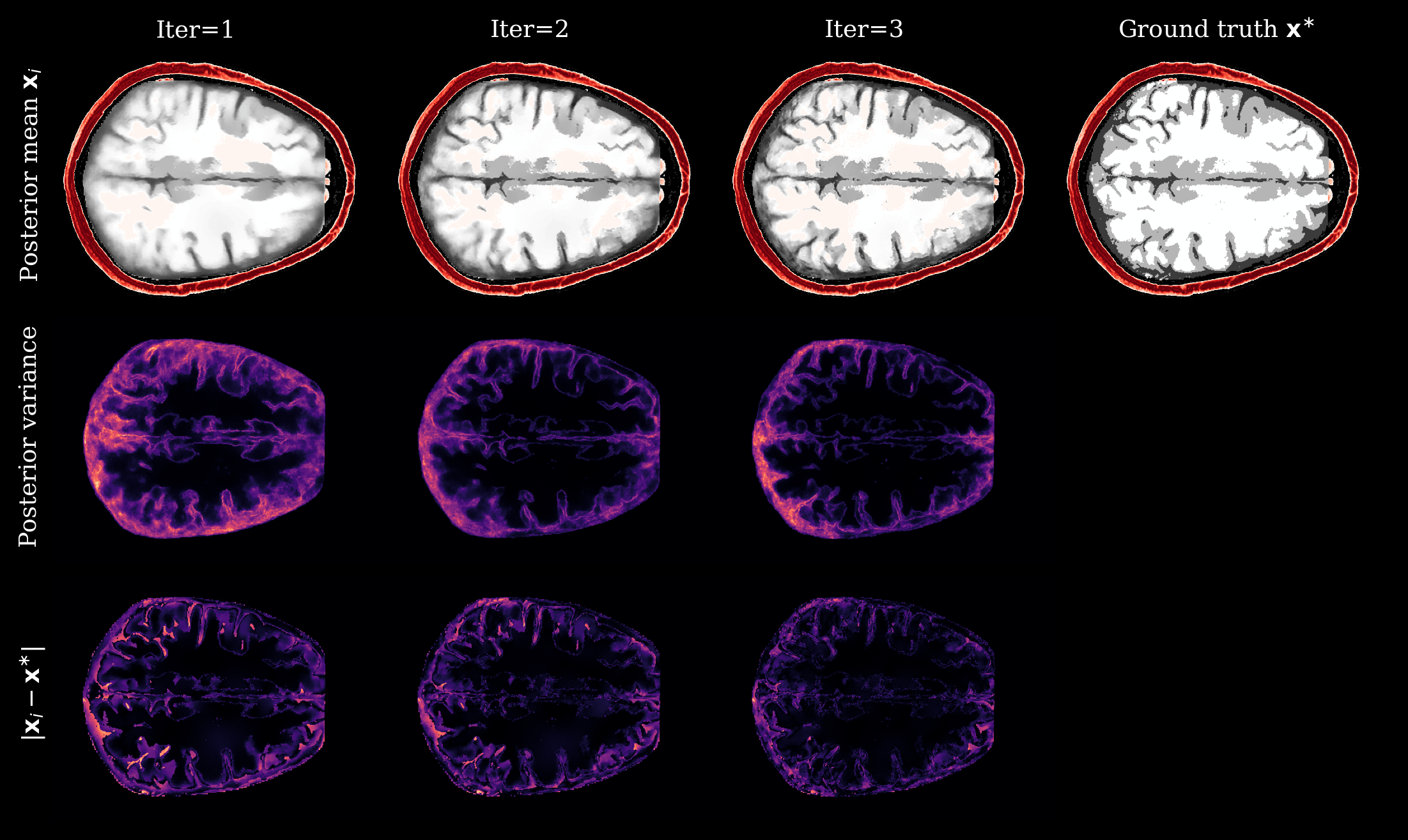}}
\end{figure}

After training our method for $L=3$ iterations, we plot posterior statistics for an unseen observation $\mathbf{y}$ in \figureref{fig:brainresult}. Each iteration improves the amortized posterior approximation as quantified by the quality of the posterior mean. Due to the receivers being limited to the top half of the simulation area (see \figureref{fig:2dsetup}) the error and uncertainty concentrates at the bottom half of the parameters. Appendix Table \ref{tab:indexes} contains average quality metrics over 200 samples from a test set. We summarize by noting that at iteration three, the average PSNR was $43.27$dB and on average the second iteration increased PSNR by $1.63$dB $\pm 0.54$dB and the third iteration increased PSNR by $2.82$dB $\pm 0.91$dB. 
%\textit{iii}) As the iterations increase, there increasing correlation between the variation of the posterior samples and the error made by the posterior mean. We will further evaluate this observation in the next section with a calibration plot. 
%RAFAEL COMMENT: COMPARE AGAINST FWI???? I GUESS FWI IS NONAMORTIZED BAYESIAN MAP WITH GAUSSIAN PRIOR ON X????
%RAFAEL COMMENT: Validate posterior using simulation based criterion. Validate its calibration got better.?

\section{Conclusion}
%\ali{conclusion could use some more work. Try to apply the ``Felix structure'' again  say something more stronger, e.g., along the lines of: we showed the first instance of X that enables Y}
\label{sec:conclusion}
We showed the first instance of a gradient-based summary statistic paired with normalizing flows that enables iteratively refining amortized posterior distributions for large dimensional and non-linear Bayesian inverse problems. Our method exchanges additional offline computation to get better amortized approximations of Bayesian posteriors. Using a Gaussian inverse problem, we showed our method converges to the ground truth posterior. We also demonstrated our method scales to a high-dimensional medical imaging problem related to expensive non-linear PDE operators.

\bibliography{sample}

\appendix

\section{Training our method}\label{apd:second}
 In this work, we follow traditional loop-unrolling schemes \citep{putzky2019invert,kelly2021recurrent} and train separate conditional normalizing flows at each step but we could also reuse the same networks or define implicit layers \cite{guan2022loop}. 
 
\begin{algorithm2e}
\caption{Training Phase}
\label{alg:train}
 % older versions of algorithm2e have \dontprintsemicolon instead
 % of the following:
 %\DontPrintSemicolon
 % older versions of algorithm2e have \linesnumbered instead of the
 % following:
 \LinesNumbered
\KwIn{Paired of observations $\{\mathbf{y}^{(n)}\}_{n=1}^{N_{train}}$ and corresponding parameters $\{\mathbf{x}^{(i)}\}_{n=1}^{N_{train}}$}
\For{$n\leftarrow 1$ \KwTo $N_{train}$}{
Generate starting fiducial: $\mathbf{ x}_0^{(n)}$\;  
$\mathbf{\Delta x}^{(n)}_0 = \mathbf{x}^{(n)} - \mathbf{ x}_0^{(n)}$\; 
Summarize observation with gradient: $\mathbf{\bar y}_0^{(n)} = \nabla_{\mathbf{x}} \log p(  \mathbf{y}^{(n)}\mid\mathbf{ x}_0^{(n)})$\;
Add pairs to dataset: $\mathcal{D}^{(n)}_0 = (\mathbf{\Delta x}_0^{(n)},\mathbf{\bar y}_0^{(n)})$ \;
}
\For{$j= 0$ \KwTo $L$}{
  
  \While{conditional normalizing flow $f_{\theta_j}$ is not converged}{
 Evaluate objective function on dataset $\mathcal{D}_j$ using \equationref{eq:train-cond-summary}\;
 Update $\theta_j$ with backpropagation\;
   } 
   \For{$n= 1$ \KwTo $N_{train}$}{
        
        Update current guess: $\mathbf{x}_{j+1}^{(n)} = \mathbf{x}_j^{(n)} + \mathbb{E}\left[\,p_{\hat \theta_j}(\mathbf{\Delta x}\mid\mathbf{\bar y}_{j}^{(n)})\right]$\;
        $\mathbf{\Delta x}^{(n)}_{j+1} = \mathbf{x}^{(n)}-\mathbf{x}_{j+1}^{(n)}$\; 
        Re-summarize observation with gradient: $\mathbf{\bar y}_{j+1}^{(n)} = \nabla_\mathbf{ x} \log p(  \mathbf{y}^{(n)}\mid\mathbf{ x}_{j+1}^{(n)})$\;

        Add pairs to dataset: $\mathcal{D}^{(n)}_{j+1} = (\mathbf{\Delta x}^{(n)}_{j+1},\mathbf{\bar y}_{j+1}^{(n)})$ \;
    }

}

\KwOut{$\hat \theta_0,\hat \theta_1, \ldots  ,\hat \theta_L$ Posterior Networks}
\end{algorithm2e}

\section{Sampling from conditional normalizing flow}\label{apd:first}

To estimate the conditional expectation $\mathbb{E}_{p_{\hat \theta}(\mathbf{\Delta x} \mid \mathbf{\bar y})}\, \left[\mathbf{\Delta x} \mid \mathbf{\bar y} \right]$ we take the empirical mean over $N_{s}$ samples. To calculate each conditional sample we pass newly sampled Gaussian noise through the inverse normalizing flow conditioned on $\mathbf{\bar y}$
\begin{equation}
\label{eq:expectation}
   \mathbb{E}_{p_{\hat \theta}(\mathbf{\Delta x} \mid \mathbf{\bar y})}\, \left[\mathbf{\Delta x} \mid \mathbf{\bar y} \right]  \approx \frac{1}{N_{\text{s}}}\sum_{i=1}^{N_{\text{s}}} \mathbf{\Delta x}^{(n)} \,\,\, \text{where}\,\, \mathbf{\Delta x}^{(n)}=f^{-1}_{\hat \theta}(\mathbf{z}^{(n)};\mathbf{\bar y}) \,\, \text{and}\,\,\mathbf{z}^{(n)}\sim \mathcal{N}(0,\,I).
\end{equation}

\section{Low training sample improvement}\label{apd:lowtrain}
 As discussed in \sectionref{sec:avi}, the main source of approximation error in amortized VI is from estimating the expectation over the joint distribution $p(\mathbf{x},\mathbf{y})$ in \equationref{eq:KL} with limited training samples. To evaluate whether our method can reduce estimation errors due to few training samples, we train the conditional normalizing flow with differing training dataset sizes: $N=400, N=1000, N=2000$. As expected, increasing the amount of training data improves the approximation as shown by the black link in \figureref{fig:improveVI}, but we can also achieve similar improvement by using our iterative method (compare black line with red line). We emphasize that traditional amortized posterior inference would be limited to the black line in \figureref{fig:improveVI} and more training data would be needed to improve the quality of the approximation whereas our method can provide better posterior approximations without additional training data.

\begin{figure}[htbp]
\floatconts
  {fig:improveVI}
  {\caption{Approximation errors due to small training datasets can be ameliorated with our iterative method.}}
  {%
   \subfigure[Posterior mean][b]{\label{fig:improvemean}%
      \includegraphics[width=0.3\linewidth]{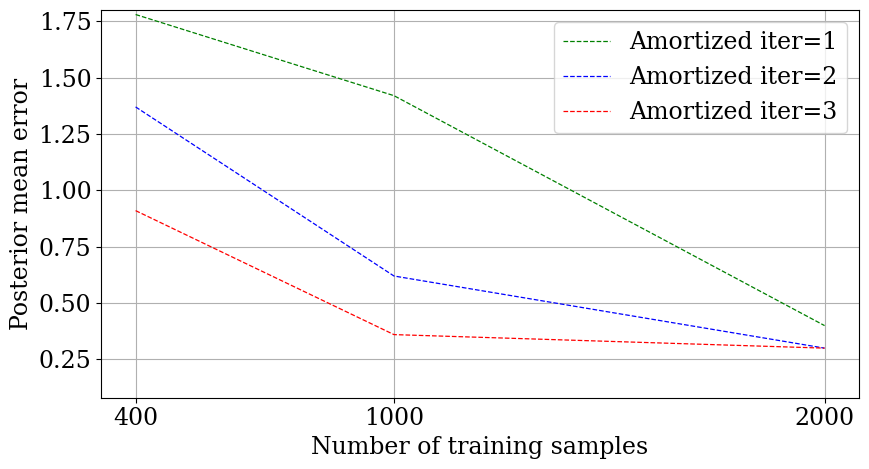}} 
    %\hspace{0.3\textwidth}
    \subfigure[Posterior covariance][b]{\label{fig:improvecov}%
      \includegraphics[width=0.3\linewidth]{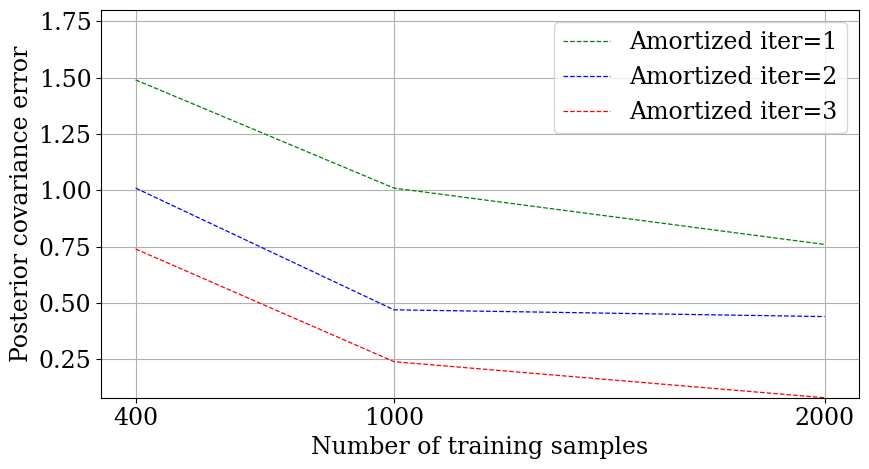}}%
  }
\end{figure}

\section{Average performance for transcranial ultrasound}\label{apd:performance}

For $200$ samples from an unseen test set, we simulate observations $\mathbf{y} $ and calculate the posterior at each iteration using our method. The average metrics for Peak Signal Noise Ratio (PSNR), Structural Similarity Index Measure (SSIM) and Root Mean Squared Error (RMSE) are shown in Table \ref{tab:indexes}. 

\begin{table}
 \begin{center}
  \begin{tabular}{lccccc}
    \toprule
    \multirow{1}{*}{Posterior mean} &
      \multicolumn{1}{c}{PSNR $\uparrow$} & \multicolumn{1}{c}{SSIM $\uparrow$} & \multicolumn{1}{c}{RMSE $\downarrow$} \\
      \midrule
    Non-iterative  ($\mathbf{x}_1$)   & 40.45  & 0.977  & 0.00964    \\
    Ours iter=2 \, ($\mathbf{x}_2$)  & 42.08  &  0.982  & 0.00806 \\
    Ours iter=3 \, ($\mathbf{x}_3$) & \textbf{43.27}  & \textbf{0.985}  & \textbf{0.00708} \\
    \bottomrule
  \end{tabular}
   \end{center}
    \caption{Image reconstruction quality metric comparison} 
     \label{tab:indexes}
\end{table}

\end{document}